\documentclass[letterpaper,conference]{ieeeconf}
\pdfoutput=1
\usepackage{amsmath}
\usepackage{cleveref}
\usepackage{makecell}
\usepackage{afterpage}
\usepackage{gensymb}
\usepackage{paralist}
\usepackage{multirow}
\usepackage{booktabs}
\usepackage{graphicx}
\usepackage{url}
\usepackage{float}
\usepackage{censor}

\renewcommand{\baselinestretch}{0.98} 

\begin{document}


\title{Characterizing Wildlife Response to Biomimetic and Conventional Underwater Vehicles}

\author{Huy Pham$^{1*}$, Levi Cai$^{2,3*}$, Yogesh Girdhar$^{3\dagger}$, Daniela Rus$^{4\dagger}$, Zach J. Patterson$^{1,4\dagger}$
\thanks{$^{1}$ Mechanical and Aerospace Engineering, Case Western Reserve University.}%
\thanks{$^{2}$ Computer Science and ESIIL, University of Colorado at Boulder.}%
\thanks{$^{3}$ Applied Ocean Physics and Engineering Department, Woods Hole Oceanographic Institution.}%
\thanks{$^{4}$ Computer Science and Artificial Intelligence Laboratory, Massachusetts Institute of Technology.}%
\thanks{$^{*}$ Equal Contribution, $^{\dagger}$ Equal Advising}%
\thanks{This work has been submitted to the IEEE for possible publication. Copyright may be transferred without notice, after which this version may no longer be accessible}%
}

\maketitle

\begin{abstract}
 Autonomous underwater vehicles (AUVs) are a promising alternative to divers for scalable collection of natural ocean ecology data. However, these robots may disturb local fauna and cause drastic behavioral differences compared to other monitoring techniques, decreasing their value as scientific tools. A promising prospect is to make AUVs that are more biomimetic, with the hope that taking on the form and behavior of a non-predatory animal may reduce adverse responses. We present the first dataset comparing fish disturbance in response to a conventional thruster-driven AUV, a sea turtle inspired flipper-driven AUV, and a diver. Experiments were conducted at a biodiversity hotspot in a Caribbean reef, and images of the scene were analyzed using computer vision to localize and study fish behavior change. Both AUVs cause measurable changes in fish behavior. Although no between-robot differences remain significant after correction for multiple comparisons, point estimates generally favor the biomimetic AUV, motivating larger studies capable of resolving modest effects and further design changes to optimize for disturbance. We also observe larger responses during diver transects than during AUV transects, although this exploratory comparison is based on a small diver sample with several confounds. While conclusions must be taken as preliminary due to operational and experimental limitations, this study provides the community with a first known dataset and benchmarks to quantify the behavioral impacts of biomimetic robots for ecological monitoring in the wild.
\end{abstract}

\section{Introduction}
The ocean covers 71\% of the earth's surface and is critically important for the function of the biosphere and global economy \cite{costanza1999ecological}. Yet, the sheer size of the ocean makes survey of targeted aquatic ecosystems difficult to achieve at scale. Conventional torpedo-shaped long range autonomous underwater vehicles (AUVs) can collect data autonomously over long distances, but cannot enter complex marine environments such as reefs, where natural life is abundant and critical for the maintenance of ocean resources \cite{sahoo2019advancements}. Therefore these environments have historically been studied by divers. While this has been an invaluable source of scientific insight, diver-based observation is inherently limited by depth, duration, and safety constraints. Additionally, the scale of the approach is limited by the availability of scientific divers, which inherently limits the amount of sites that can be explored. Furthermore, thanks to the ability to carry and respond to a wide variety of sensors in realtime, robots can unlock new possibilities in exploration strategy that enables the capture of different types and quality of information \cite{mccammon2026autonomous, powell_first_2022, clarke_using_2009}. Thus, AUVs with the agility to enter a coral reef and perform scientific monitoring of the local ecosystem offer the possibility of a valuable tool for ocean science. 


\begin{figure}
    \centering
    \includegraphics[width=0.95\linewidth]{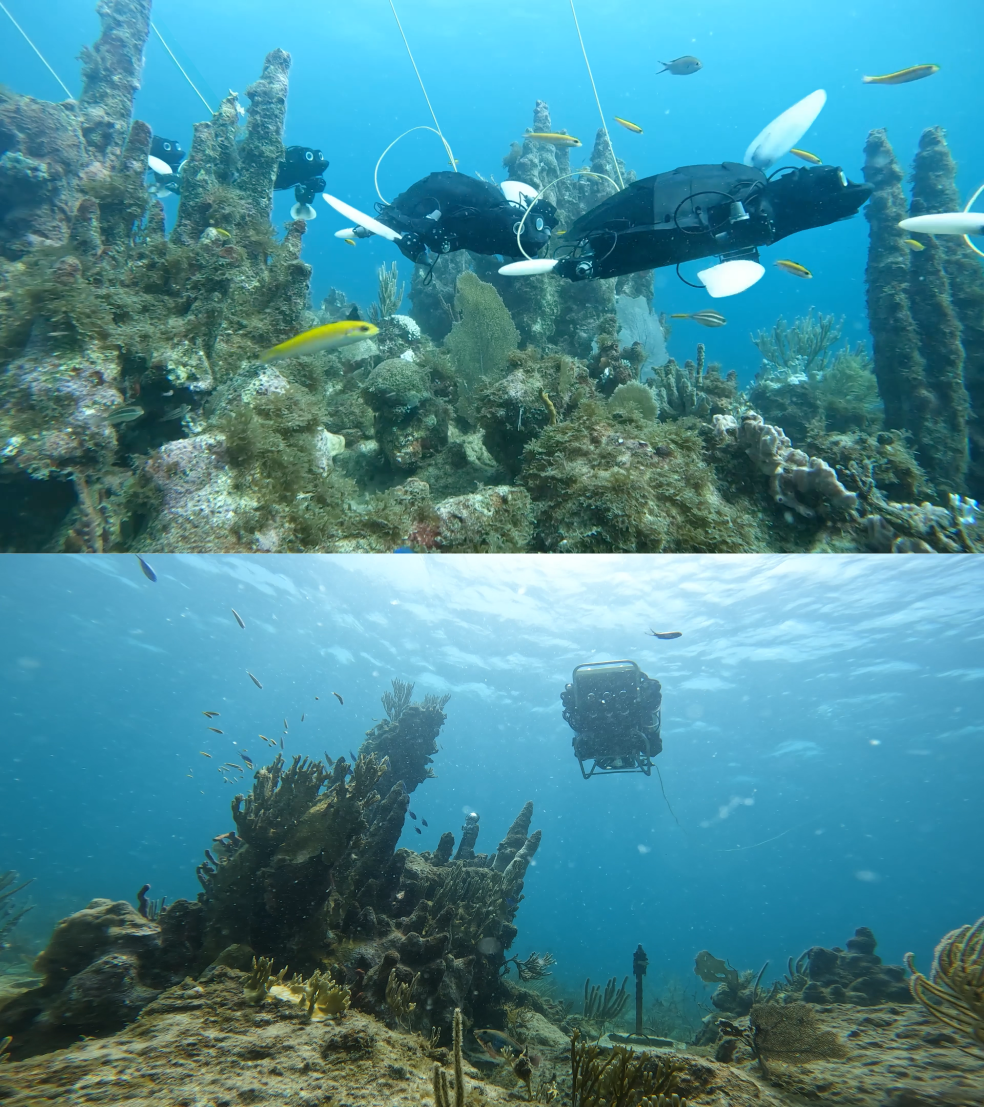}
    \caption{We characterize the response of wild fish in a Caribbean coral reef to a biomimetic sea turtle robot (top) and a conventional thruster-driven AUV (bottom). Specifically, we seek to quantify the observer bias of each platform in comparison to a human diver by quantifying the ``disturbance'' created by each robot's presence.}
    \label{fig:bots}
\end{figure}

One potential concern for the working ocean scientist on the use of robots is the potential to introduce observer bias into the marine habitat through disturbance of the local fauna. If the presence of a robot changes the way that the ecosystem behaves, its value as a scientific instrument diminishes. Several previous studies have shown bias induced by the presence of robots in small tanks \cite{landgraf_socially_2020, musiolek_robofish_2020, swain_real-time_2012} and in aquaculture settings \cite{kruusmaa_salmon_2020}, with more extensive studies on size and shape impacts in \cite{zhang_avoidance_2025, zhang_farmed_2024}. In the field, there have been a few attempts to study fish response to AUVs \cite{campbell_assessment_2021, stoner_evaluating_2008, benoit-bird_schrodingers_2023, cai2025measuring}, and have similarly found that bias does exist. 

One proposed approach to reduce the disturbance of AUVs is to produce biomimetic systems that look and act more like ocean wildlife. This is frequently cited as a motivation for basic research on such aquatic platforms \cite{fish2020advantages}, but it has rarely been examined in field settings. However, studies such as \cite{asunsolo-rivera2023behaviour} have shown promise that biomimetic non-predatory forms may reduce certain responses. Additionally, the aforementioned in-tank and aquaculture experiments have shown that conspecific or turtle-like robots can induce different behaviors such as attraction and more neutral responses, respectively. However, to our knowledge, no such study has been conducted on organisms in the wild, and have not been directly compared between traditional vehicles, biomimetic vehicles, natural predators, and human divers.


In this work, we seek to compare the disturbance created by a biomimetic sea turtle robot (Fig. \ref{fig:bots} top) to that of a conventional thruster-driven AUV (Fig. \ref{fig:bots} bottom) and to a human diver by performing low altitude transects through a coral reef environment in the US Virgin Islands. Using fixed cameras placed on the seabed, we estimate the disturbance of each survey method by labeling and localizing fish in the recorded videos.

\begin{itemize}
    \item The first comparative deployment of a conventional and biomimetic AUV.
    \item Data on fish response to different AUVs and to a diver.
    \item An open source pipeline to analyze such data for fish disturbance\footnote{\url{https://github.com/CyPhiLab/turtle_dm}}.
    \item The first evidence that AUVs may be less intrusive than divers, and that biomimetic AUVs may be less intrusive than conventional ones.
\end{itemize}

\section{Methods}\label{sec:deploy}
\subsection{Protocol}
Measuring wild animal responses in natural environments is a challenging task. There is no access to laboratory-grade motion capture, tagging, environment control, etc. Combined with the operational complexity of deploying experimental robots, this requires compromises in the data collection method as well as the type of data that it is possible to collect. Thus, while multimodal sensing has been shown to be important for characterizing biodiversity in ocean environments \cite{mccammon2026autonomous}, we believe that simpler data and protocol is a better choice at this stage for comparing robot disturbance. We therefore choose to monitor disturbance with 3rd person, fixed cameras. While ideally we can eventually move towards using robot cameras for this purpose, fixed cameras allow us to more easily measure baseline behaviors of fish in addition to how they respond to a robot. By centering the analysis around images and using course-grained, high level metrics, data collection is simplified, requiring only a single well-placed camera. 

In order to measure disturbance for a given entity (i.e. robots, divers, animals, etc), we use the following experimental protocol. We place a desired number of cameras facing a coral head that serves as a reef ``hotspot,'' ensuring that the cameras are placed to see both the vehicle transect and hotspot as best as possible. We also place AprilTags near the chosen coral head and at waypoints some distance from the coral head on either side, forming a transect route of three points. The AprilTags can be used as landmarks to guide the pilot during operation, but eventually they could also be used to perform automated localization and control. After waiting a few minutes (allowing fish to rehabituate from the potential disturbance of camera placement), a robot is deployed. The robot performs transects along the route designated by the AprilTags. Each transect proceeds as follows. First, the robot dwells at the first tag for a control period. Then, the robot swims through the reef and over the center AprilTag near the primary coral head, stopping to dwell over the third AprilTag. The robot again idles for the control period before performing another transect in the reverse direction. This process repeats until the block of experiments is complete.

Once experiments are complete, the collected video data is analyzed. Fish are labeled, either by hand or computer vision. We note that it is important to control for different days and times of day, or even behaviors across different sites (as baseline behaviors and species-present may be different). Our comparison protocol utilizes the \textit{change} from baseline (measured during the control points of the transect) induced as the robot passes the coral head. We cannot guarantee that looking at changes in behavior eliminates bias with respect to the aforementioned factors, but it should provide a good start. 

\subsection{Robots}
Both robots' designs are described in other work. The thruster-driven AUV, CUREE, uses six Blue Robotics T200 thrusters to achieve six degrees of freedom and is outfitted with a variety of sensors including several cameras for surveying and a Doppler Velocity Logger (DVL) for localization and altitude estimation  \cite{girdhar2023curee}. CUREE can operate autonomously or via remote teleoperation. The biomimetic robot, Crush, has the form of a sea turtle \cite{patterson2026autonomous}. It generates thrust with a pair of front flippers and uses a pair of rear flippers as rudders for steering and trim. The flippers were injection molded out of silicone rubber (DragonSkin 30). All four limbs are driven by DYNAMIXEL XW540-T260-R servomotors. We found that the motor connectors were not sufficiently waterproof to prevent small amounts of salt water from entering, which caused rapid motor failure. After sealing the connectors with epoxy, we encountered no further motor issues during our deployments. The robot has a depth sensor and two cameras (Deepwater Exploration exploreHD) placed into the eye sockets.

\subsection{Experiments}
The robots were deployed in November 2024 at the Yawzi coral reef site off of the coast of St. John in the US Virgin Islands. The experimental protocol was inspired by previous work. Both robots were teleoperated in order to ensure desired behavior and prevent collisions with the reef. CUREE has altitude control and was set to hover at an altitude of 1.5 meters, which corresponded to a depth of about 7 meters at the reef site. Crush has depth control, but it was not used during the study due to mechanical buoyancy issues (which will be further discussed in study limitations). Instead, the operator manually controlled the robot's depth. Two GoPro cameras were placed by divers on the benthic surface at opposite ends of the reef, facing the primary coral heads. These cameras record video that is later used to monitor fish behavior from a fixed perspective.

Transects of each robot are attempted as follows. The robot loiters at one end of the reef for 30-plus seconds and swims through the reef before loitering at the other end of the reef for 30-plus seconds. This process is repeated until experiments are completed. We collected $N=17$ turtle robot transects and $N=18$ CUREE robot transects. As a baseline, on the previous day, we also collected $N=4$ diver transects. Finally, during time in between experiments, a shark swam through the reef. Although it is only a single sample, we report it as an interesting comparison. While we were able to precisely control CUREE during this process, we had some issues controlling the turtle robot in the strong currents and thus transects were not always ideal. On the day of deployment, the robots were tested for their entire set of transects, one after the other. In other words, we first deployed the turtle robot and completed experiments with it, then deployed CUREE and completed experiments with it. This methodology introduces obvious limitations to the dataset, but it was unavoidable on this particular trip due to time constraints and the operational difficulty of deploying both robots at the same time. We elaborate on this and other limitations in the Discussion.

\begin{figure}
    \centering
    \includegraphics[width=0.95\linewidth]{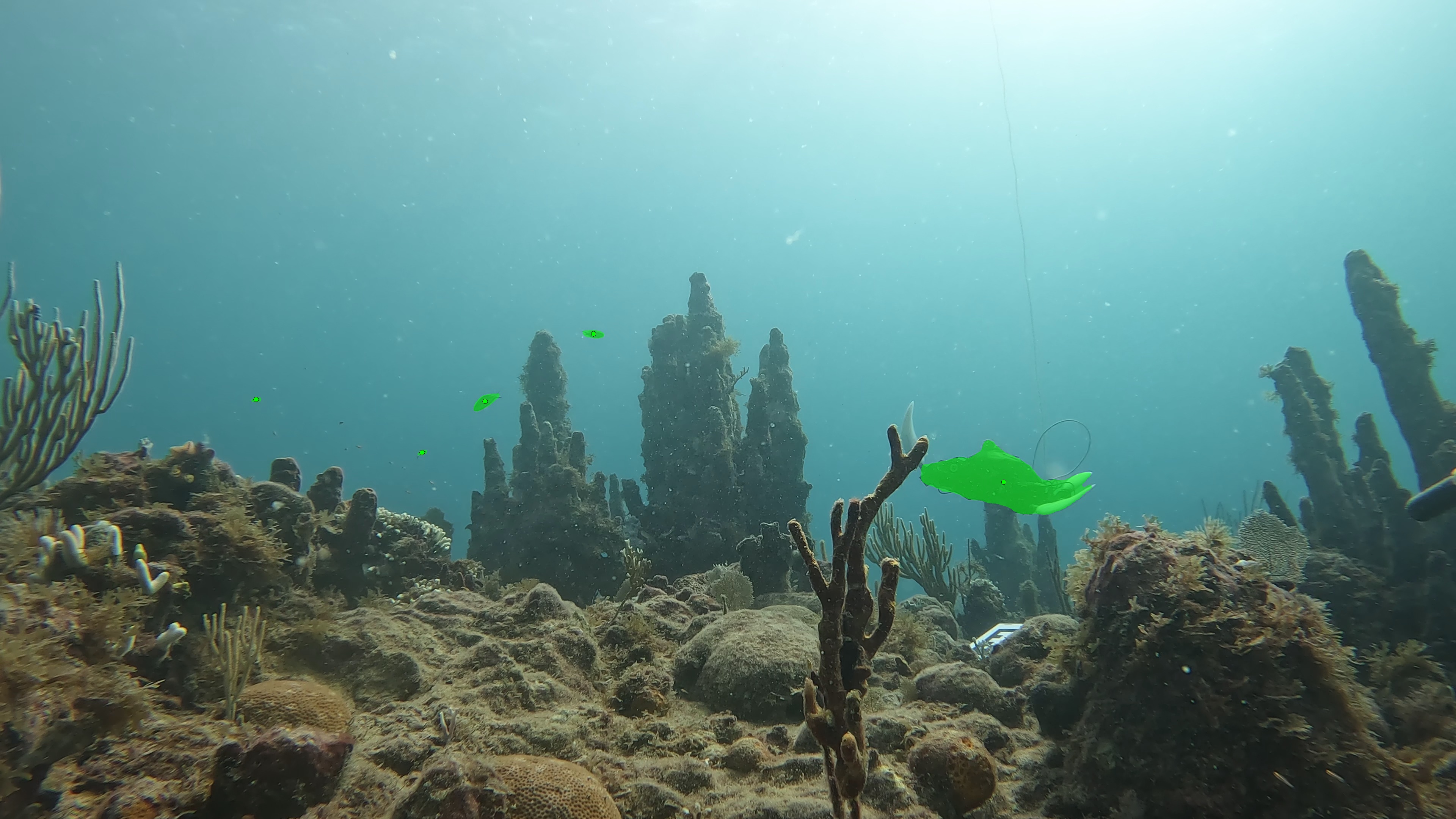}
    \includegraphics[width=0.95\linewidth]{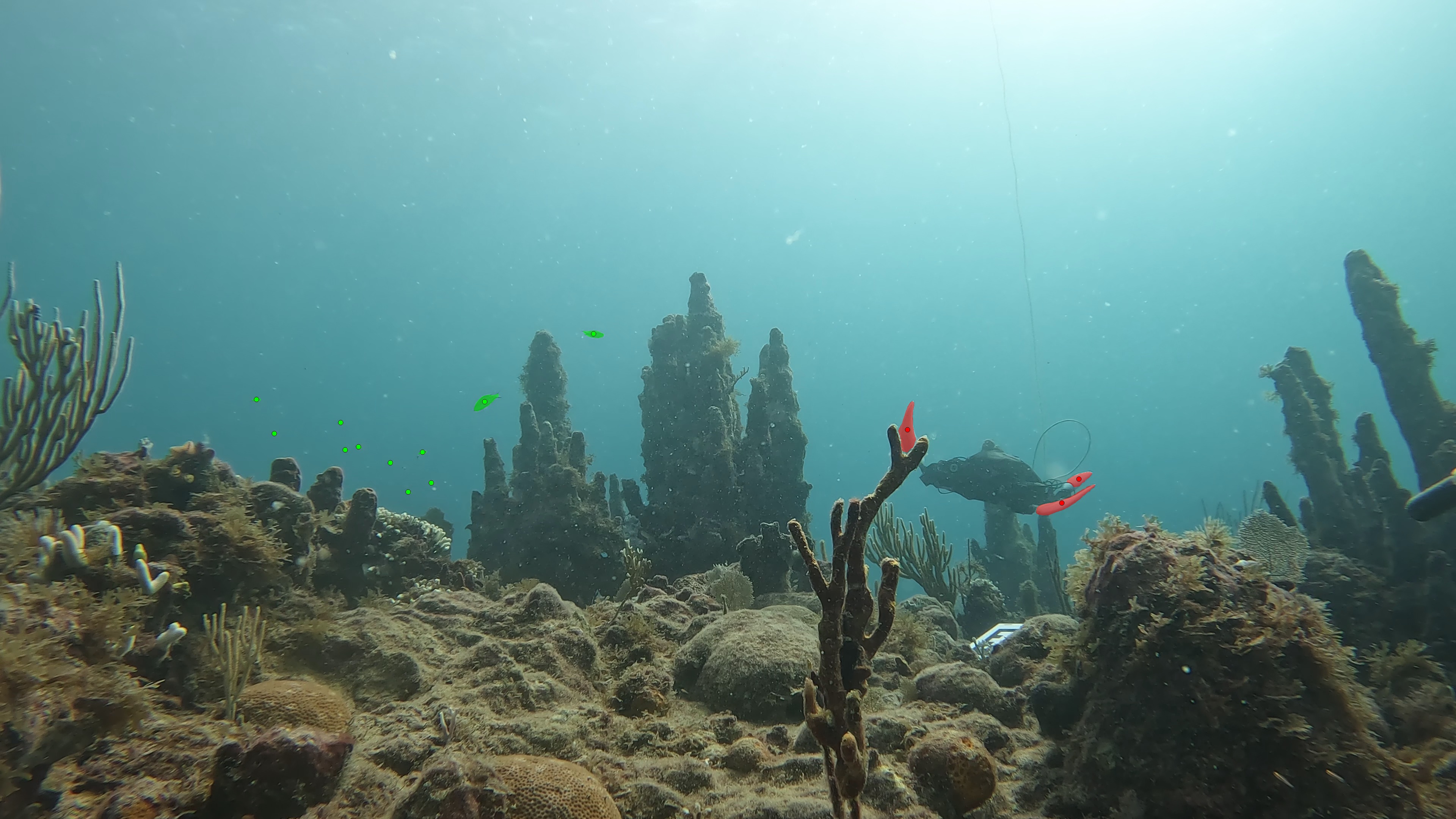}
    \caption{A comparison of performance between the original pretrained (top) and fine-tuned (bottom) SAM3 models. Fish and turtle flippers are segmented in green and red, respectively. It can be seen that before fine-tuning, the model confuses the turtle and its flippers as a fish, and misses small fish. The fine-tuned model significantly improves performance.}
    \label{fig:finetuning}
\end{figure}
\subsection{Identifying fish in images}




To automate the fish detection process, we utilize Segment Anything Model (SAM) 3 \cite{carion2025sam3}, a foundation computer vision model for detecting and segmenting multiple objects based on concept prompts. Initial attempts with the baseline SAM3 model on our videos showed poor results because the original model is not capable of capturing fish from far distances under dim underwater lighting, and often confuses Crush and its flippers as fish. To address these limitations, we fine-tuned SAM3 tailored for our fish monitoring task, improving detection accuracy for both fish and turtle flippers in the challenging underwater environment.

We extracted 80 frames that contain fish and turtle flippers from the video footage to collect training data. Fish and turtle flippers were then manually segmented using Label Studio  \cite{tkachenko2020label}, resulting in 1,801 total annotations across all categories. The frames were split into training and validation sets with an 80/20 ratio, and the split was performed deterministically to ensure reproducibility.  Consequently, the fish category comprises 1,271 training annotations and 327 validation annotations, while the turtle flipper category comprises 154 training annotations and 49 validation annotations. Finetuning was conducted over 40 epochs with validation performed every 5 epochs. 

Training validation is performed using the Common Objects in Context (COCO) evaluation framework \cite{lin2015microsoftcococommonobjects}. COCO computes Average Precision (AP) and Average Recall (AR) over ten Intersection over Union (IoU) thresholds from 0.50 to 0.95 in increments of 0.05. Results are reported for three object-size categories following the standard COCO definitions: small ($area <  1024~{px}^2$), medium ($1024 \leq area < 9216~{px}^2$), and large ($area \geq  9216~{px}^2$). Validations on our fine-tuned model reveal high performance on large objects, yielding an AP of 0.966 and an AR of 0.971. Performance remains robust for medium-sized objects, with an AP of 0.717 and an AR of 0.757. However, a noticeable performance bottleneck occurs within the small object category, where AP drops to 0.372 and AR to 0.408. Very small fish were often observed near the coral structure, but we did not quantify how missed detections affected the disturbance metrics. Because hiding by small fish may itself be a disturbance response, size-dependent detection performance remains a limitation of this methodology.


\subsection{Disturbance metrics}
After producing the annotated dataset, we explored a number of candidate metrics for analyzing the disturbance of the fish. Prior literature examined the \textbf{shelter distance}, or the euclidean distance from an arbitrary point placed at the center of the primary coral head in the scene. The assumption in using this metric is that the fish seek shelter when disturbed. In our data, we found that the fish had alternative flight responses in addition to this one: most notably, they frequently fled upwards and outwards, away from the reef altogether. Therefore, we found that this metric did a poor job of capturing visually obvious disturbance and we examined other candidate metrics as well.

The most informative signal was found to be \textbf{fish count}, for which we simply count the number of tagged fish in the scene at each timestep. We also examined \textbf{vertical position}, or the mean vertical position of the fish in the scene in pixels, which is an alternative spatial metric to the shelter distance. The motivation for looking at this variable was the observation that fish often fled upwards in response to the transects. \textbf{Centroid speed} is the frame-to-frame movement of the school’s center. At each frame we take all detected fish positions, compute their mean location to get the centroid, then compare that centroid to the centroid from the previous frame. The Euclidean distance between those two centroids is the centroid speed for that frame, in px/frame. The \textbf{nearest neighbor distance}, or the minimum value of the pairwise euclidean distance between objects, has been used in the literature to characterize the behavior of shoaling fish and captures local clustering \cite{herbert-read2011inferring}. On the other hand, \textbf{spatial entropy} quantifies the global dispersal of the fish \cite{karakaya2021acute}: low spatial entropy means the fish are concentrated in a small part of the frame while high spatial entropy means fish are spread across many parts of the frame more evenly. It is computed by getting the histogram of fish positions on an 8×8 spatial grid, converting bin counts to probabilities, and calculating $H = -\sum p_i \ln{}(p_i)$. 

\subsection{Statistical Analysis}

Fish behavioral responses to each robot were quantified using a within-transect comparison of behavior at the peak of the disturbance to pre- and post-disturbance baselines. We identify transects by manually tagging the time that the robot starts and stops moving. The time of each trial was normalized to unit progress, such that the transect start and end corresponded to approximately 0 and 1 respectively. Frames were assigned to one of three non-overlapping windows based on this normalized progress: a \emph{start} window ($[0.00, 0.25)$), a \emph{middle} window ($[0.375, 0.625]$), and an \emph{end} window ($(0.75, 1.00]$). 
The median value of each behavioral metric was computed within each window for every trial, yielding three per-trial scalars.
The change in each metrics at the center of the transect compared to the beginning and end was computed for each trial as
\begin{equation}
    \Delta = x_{\text{middle}} - \tfrac{1}{2}\!\left(x_{\text{start}} + x_{\text{end}}\right),
\end{equation}
\noindent
where $x_w$ denotes the within-window median for window $w$.

For each robot and metric, a one-sample $t$-test was used to assess whether the mean $\Delta$ differed significantly from zero
($H_0\colon \mu_\Delta = 0$) at the center of the transect compared to the ends.
The primary between-condition comparison was CUREE versus turtle robot. For each metric, a Welch two-sample $t$-test was applied to the per-transect $\Delta$ values of the two conditions.
The effect size was estimated as the difference in condition means
($\bar{\Delta}_{\text{AUV}} - \bar{\Delta}_{\text{turtle}}$), with a 95\% confidence interval derived from the Welch standard error. 
Comparisons involving the diver condition ($n = 4$) were treated as exploratory and are not subject to formal inference.
Benjamini--Hochberg false discovery rate (FDR) correction~\cite{benjamini1995controlling} was applied across the six primary AUV-versus-turtle metric tests, with a target FDR of 5\%. Within-condition tests and diver comparisons were kept separate from this correction procedure. All analyses were conducted in Python~3 using \texttt{scipy}~\cite{virtanen2020scipy}
and \texttt{statsmodels}~\cite{seabold2010statsmodels}.

\subsection{Audio analysis}
The GoPros used to record the coral reef during the experiments also have microphones, which allows the analysis of audio data to characterize the sound profiles of the two robots. The GoPros were placed on either end of the reef, and thus the average of their recorded sound approximates the audio that was present within the reef. We compare the root mean square (RMS) sound pressure. The audio signal $y[t]$ is divided into overlapping frames of length $L = 2048$ samples
with hop size $h = 512$ samples. For each frame $k$, the root-mean-square amplitude is
\begin{equation}
    \text{RMS}_k = \sqrt{\frac{1}{L} \sum_{n=0}^{L-1} y[kh + n]^2}
\end{equation}
This gives a time series of RMS values with frame centers at
$t_k = (kh + L/2)\,/\,f_s$, where $f_s$ is the sampling rate.
Each clip's time axis is centered on the transect midpoint and the resulting series is interpolated onto a common grid spanning $\pm 40$\,s. The mean and standard deviation across clips within each group are computed pointwise on this grid. Because the GoPro microphones were not calibrated, RMS values are reported in arbitrary digital units and support only relative comparisons within this recording setup; they cannot be related directly to fish hearing thresholds.


\section{Results}\label{sec:results}
\begin{figure*}
  \centering
  \includegraphics[width=0.96\linewidth]{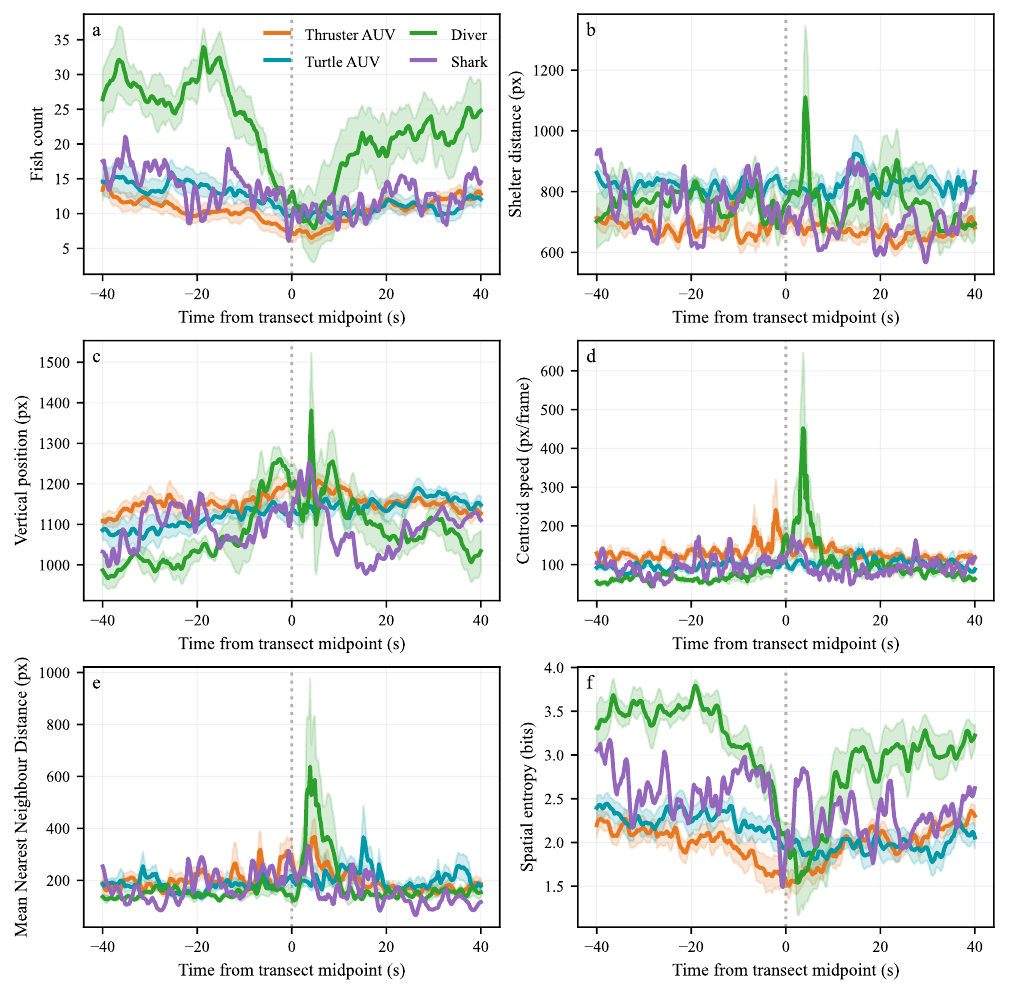}
  \caption{Mean $\pm$ SEM of six fish-assemblage metrics over time relative to the transect midpoint for each robot condition (tAUV, Turtle, Diver, Shark). Metrics shown are (\textbf{a})~fish count, (\textbf{b})~shelter distance, (\textbf{c})~vertical position, (\textbf{d})~centroid speed, (\textbf{e})~mean nearest-neighbour distance, and (\textbf{f})~spatial entropy. The dotted vertical line at $t = 0$ marks the transect midpoint (closest approach). Shaded regions indicate $\pm 1$ standard error across trials.
}
  \label{fig:metrics}
\end{figure*}
The calculated metrics for all transects are reported in Figure \ref{fig:metrics} and corresponding within-robot statistical analyses are presented in Table \ref{tab:within_condition}. We remind the reader that $N=17$ for the turtle robot, $N=18$ for the thruster-driven AUV, and $N=4$ for the diver. The diver transects were also collected on the preceding day, when baseline fish counts were approximately 25--30 compared with 10--15 during the robot experiments. Thus, day-specific conditions, time, or species composition may contribute to the diver comparison. Qualitatively, the most immediate conclusions are that all conditions cause a characteristic dip in the fish count at the center of the transect, which then recovers towards the pre-disturbance condition. This classic profile indicates that all conditions disturb the fish. Looking at the plot, the thruster AUV sees a slightly larger drop in \textbf{fish count} ($-3.12$) than the turtle robot ($-2.56$), although the difference was not statistically significant ($p=0.49$). From its higher baseline, the diver has a larger drop ($-9.76$). \textbf{Shelter distance}, which was previously used as a signal of disturbance in related work \cite{cai2025measuring}, displays no notable trends. The thruster AUV and diver both see an increase in \textbf{vertical position} on average ($37.58$, $p<0.05$ and $157.3$, $p<0.05$ respectively). The turtle's response is a bit different in that it continues rising during and after the transect before beginning a recovery to baseline later, driving the mean change to zero. However, due to the large variance in this metric, differences between the turtle and thruster AUV are not statistically significant. 

\begin{table}
\caption{Per-robot response difference during disturbance.}
\label{tab:within_condition}
\begin{tabular}{llrrr}
\toprule
Robot & Metric & $\Delta$ & 95\% CI & $p$ \\
\midrule
\multirow[c]{6}{*}{\makecell{Thruster\\AUV}} & Fish count & -3.12 & [-4.38, -1.87] & $<0.001$ \\
 & Spatial entropy & -0.46 & [-0.65, -0.27] & $<0.001$ \\
 & Centroid speed & 21.50 & [6.51, 36.48] & 0.008 \\
 & Vertical position & 37.58 & [2.64, 72.51] & 0.037 \\
 & Mean NND & 47.80 & [-2.21, 97.82] & 0.060 \\
 & Shelter distance & -3.46 & [-46.83, 39.91] & 0.868 \\
\multirow[c]{6}{*}{Diver} & Fish count & -9.76 & [-12.61, -6.91] & 0.002 \\
 & Spatial entropy & -0.85 & [-1.11, -0.58] & 0.002 \\
 & Vertical position & 157.30 & [71.73, 242.86] & 0.010 \\
 & Centroid speed & 33.06 & [-29.29, 95.42] & 0.190 \\
 & Mean NND & 35.80 & [-79.38, 150.97] & 0.396 \\
 & Shelter distance & 42.06 & [-191.86, 275.97] & 0.607 \\
\multirow[c]{6}{*}{\makecell{Turtle\\AUV}} & Fish count & -2.56 & [-3.69, -1.43] & $<0.001$ \\
 & Spatial entropy & -0.19 & [-0.37, -0.02] & 0.032 \\
 & Mean NND & 15.07 & [-6.93, 37.06] & 0.166 \\
 & Centroid speed & 7.51 & [-4.47, 19.50] & 0.202 \\
 & Shelter distance & -18.72 & [-50.73, 13.28] & 0.233 \\
 & Vertical position & 0.67 & [-33.08, 34.41] & 0.967 \\
\bottomrule
\end{tabular}
\end{table}

\textbf{Centroid speed} shows a significant increase for the thruster AUV ($+21.50$, $p=0.008$), suggesting increased group-level movement during the disturbance. The turtle robot shows a similar but weaker and non-significant trend ($+7.51$, $p=0.20$), while the diver shows a large but highly variable increase ($+33.06$, $p=0.19$). The difference between thruster AUV and turtle does not reach significance ($p=0.13$). \textbf{Mean nearest-neighbour distance} was not significantly different on average for any of the conditions.
The most consistent signal beyond fish count is \textbf{spatial entropy}, which decreases significantly for both the thruster AUV ($-0.46$ bits, $p<0.001$) and turtle robot ($-0.19$ bits, $p=0.032$), and most strongly for the diver ($-0.85$ bits, $p=0.002$). This indicates that fish become more spatially concentrated as the disturbance passes through, consistent with a shoaling or refuge-seeking response.

Between-robot comparisons were also performed.
No metric showed a statistically significant difference between the thruster AUV and turtle robot after Benjamini--Hochberg correction. The largest uncorrected effect was for spatial entropy (difference $= -0.27$, $p = 0.037$, $p_\text{FDR} = 0.22$), suggesting the thruster AUV may elicit a stronger spatial concentration response than the turtle robot, though this result does not survive correction for multiple comparisons. We note that with $N = 17$--$18$ transects per condition the study is likely underpowered to detect moderate between-robot differences, as reflected by the large confidence intervals. 

\begin{figure*}
    \centering
    \includegraphics[width=0.99\linewidth]{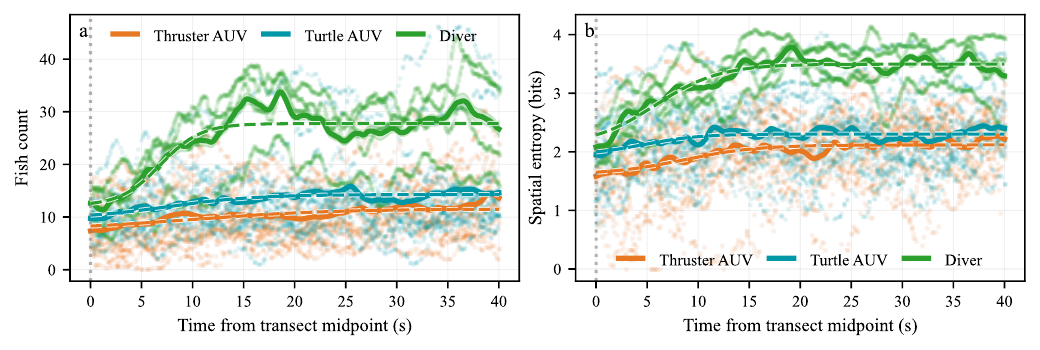}
    \caption{Flight initiation distance proxied by time (0~s = closest approach, 40~s = far from robot) for (\textbf{a})~fish count and (\textbf{b})~spatial entropy. Points show individual per-frame metrics (subsampled for clarity), solid lines show binned mean~$\pm$~standard error, dashed lines show the fitted logistic model.}
    \label{fig:fid}
\end{figure*}

The single shark passage provides an interesting natural-predator reference in Fig. \ref{fig:metrics}, with a visible response across several metrics. However, one opportunistic event cannot characterize a typical shark response or support uncertainty estimates, so we retain it as a descriptive comparison but exclude it from the statistical tests and flight-initiation fits.

Following prior literature, we also evaluated the flight initiation distance based on the fish count and spatial entropy metrics. For these experiments, we do not have access to distances so we use time as a proxy as justified in prior study \cite{cai2025measuring}. For each robot--metric combination we modeled the response of the fish assemblage as a function of time leading up to the transect $x = |t|$, where $t$ is the time (in seconds) relative to the transect midpoint ($x = 0$).
We fitted the two-parameter logistic profile
\begin{equation}
    y(x) = L_d + \frac{L_0 - L_d}{1 + e^{-k\,(x - x_0)}}
    \label{eq:fid_logistic}
\end{equation}
to the pooled raw observations. The free parameters are
the steepness $k$ and the sigmoid inflection point $x_0$. Initial
values were set to $k = 0.2$ and $x_0 = 20$. The baseline level $L_0$ was estimated as the mean metric value over
frames with $x > 30\,\text{s}$ (far tail, near-undisturbed), and the
disturbance level $L_d$ was estimated as the mean over frames with
$x < 5\,\text{s}$ (near the midpoint, maximum disturbance).  These
values were computed separately for each robot condition using all
pooled raw per-frame observations.

The flight-initiation distance $x_{\text{FID}}$ was defined as the time at which the modeled response first reaches a fraction
$\alpha = 0.95$ of the baseline level, i.e.
\begin{equation}
    \frac{L_0 - L_d}{1 + e^{-k\,(x_{\text{FID}} - x_0)}} + L_d
    = \alpha\,L_0.
    \label{eq:xfid}
\end{equation}
Rearranging gives the closed-form solution
\begin{equation}
    x_{\text{FID}} = x_0 - \frac{1}{k}\ln\!\left(
        \frac{L_0\,(1-\alpha)}{\alpha\,L_0 - L_d}
    \right).
    \label{eq:xfid_closed}
\end{equation}
$x_{\text{FID}}$ was reported as undefined when the argument of the
logarithm was non-positive (i.e.\ when the modeled response does not reach the $\alpha$ threshold within the observed range). Results of the fit are plotted in Fig. \ref{fig:fid} and the calculated FIDs are summarized as follows. For the fish count metric, $x_{tAUV}=22.02$s, $x_{turtle}=14.80$s, $x_{diver}=11.43$s. For the spatial entropy metric, $x_{tAUV}=13.37$s, $x_{turtle}=6.73$s, $x_{diver}=11.94$s.

Finally, we report the relative sound magnitude of the thruster AUV, the turtle robot, and a control with no robot present. We find that over the course of each trial, the average RMS amplitude in arbitrary digital units was $0.0072 \pm 0.0021$ for the thruster AUV, $0.0035 \pm 0.0011$ for the turtle, and $0.0032 \pm 0.0019$ for no robot. At the center of the transect, over a window from -5 seconds to 5 seconds from transect midpoint, the average RMS amplitude was $0.0110 \pm 0.0040$ for the thruster AUV, $0.0042 \pm 0.0031$ for the turtle, and $0.0037 \pm 0.0028$ for no robot.

\begin{figure}
    \centering
    \includegraphics[width=0.99\linewidth]{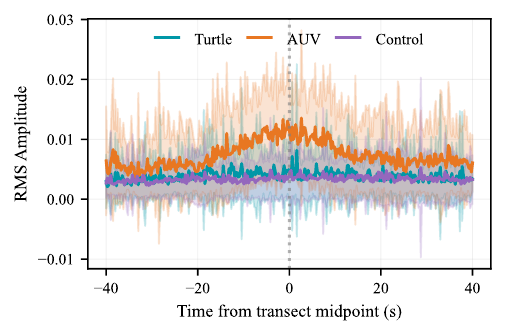}
    \caption{Relative RMS sound amplitude in arbitrary digital units from uncalibrated GoPro microphones.}
    \label{fig:rms}
\end{figure}



\section{Discussion}

Before discussing caveats borne out of the practical compromises that were made for our experimental design, we first offer tentative interpretations of our results. The fish response associated with the diver transects is larger in magnitude than for the robots, which both elicit similar response profiles. However, the diver sample comprised only four transects collected on a different day with substantially higher baseline fish abundance. The observed contrast may therefore reflect day-specific conditions or species composition as well as observer type, motivating a controlled, interleaved comparison.

While we did not find significant differences between the change in response induced by our robots, it is clear from Fig. \ref{fig:metrics} that the response profiles are different. Further experiments are necessary to determine if these differences are due to random variation, experimental factors, or true between robot differences.

As discussed, the study has limitations induced by practical challenges encountered during fieldwork. We will briefly summarize those challenges and their effect on the experiment, along with suggestions for future mitigation so that other researchers can prepare for similar factors. First, we had difficulty controlling the sea turtle robot during periods with high current, which led to rather messy transects relative to the thruster AUV. This behavioral difference may artificially bias the results. In this particular case, we believe that it is more likely that this behavior actually causes a larger disturbance than would otherwise be the case, and are optimistic that, with cleaner transects, the turtle robot disturbance can be further reduced. To address this, the turtle robot's swimming stroke needs to be improved to increase the thrust capacity, which involves optimization of the flipper and shoulder mechanism and the control gaits. The robot design was not optimized for high thrust or efficiency, so we believe that there are easy gains to be had in this area based on previous work \cite{vandergeest2023softrobotic}. 

The GoPro recordings showed higher relative RMS amplitude for CUREE than for the turtle robot and we did not establish a causal link between sound and disturbance. Moreover, spectral and temporal content may matter more than broadband magnitude; in particular, the turtle robot's periodic flipper stroke could remain biologically salient despite its lower RMS amplitude. Calibrated hydrophones and frequency-resolved analysis will be necessary to evaluate the importance of sound.

Another study limitation was experimental sequencing: we performed all experiments for the turtle robot before performing all experiments for CUREE. Therefore, we are unable to control for potential confounders such as time of day and other variables that may be effected by timing. 
Ideally, we would have preferred to run experiments while alternating the robots. Options to do this are to either have both robots in the water at the same time, or to alternate by bringing the robots back to the boat after smaller chunks of experiments. The former option would be ideal, but tether management makes this highly inadvisable due to the potential to tangle the tethers. While the robots can both operate off-tether \cite{patterson2026autonomous,girdhar2023curee}, the risk and harm of running into and damaging the fragile \textit{dendrogyra} pillar corals is too great as these corals are critical cornerstones of this ecosystem. On the other hand, because each deployment is time consuming and challenging, the difficulty of running the experiments increases with the number of times that the robot must be recalled to the boat. Thus, the best balance for future studies is to run interleaved sets of experiments (e.g. 10 transects for turtle robot, then 10 for CUREE, then repeat).

Finally, the robot data for this study were collected on a single day and at a single site. Future work should perform such comparative studies over a longer period of time and at more sites. There are a few reasons for this. First, getting the turtle robot ready for these challenging experiments occupied most of our time during the fieldwork. Second, the particular site that we chose for these experiments was favorable from an operational standpoint. The sea in this location was calmer, which helps both for the robot (less currents to fight) and for the operator (setting up and driving the robot is much easier when the boat is not being tossed about by waves). We explicitly avoided other, potentially more active coral reef hot spots in the area for this reason. By improving the sea turtle robot's ability to swim efficiently in currents, we open up the possibility of deployments in more complex and challenging ocean environments. 

Overall, because of the limitations described, we are actually encouraged that the biomimetic robot managed to perform at a similar level to the thruster-driven AUV despite systematic factors that likely harmed its performance. With a few upgrades such as custom motors to minimize noise, choosing a color other than black (which may look like the shadow of a predator), and better propulsion to operate in currents, we believe that the turtle robot's disturbance should be able to be substantially decreased from this baseline.



\section{Conclusion}\label{sec:conclusion}
In this work we presented the first analysis comparing the disturbance response of wild fish in a coral reef to a biomimetic robot and a thruster-driven AUV. We also report an exploratory comparison with a diver. Responses associated with the diver transects were larger, but the small sample and different collection day are confounders. More data are necessary to understand the observer bias induced by each survey method. Our work highlights the necessity of having hardened and very capable robot platforms for conducting this kind of fieldwork. We are continuing to make improvements to the turtle robot platform with the goal of improving its ability to blend naturally into aquatic ecosystems. 


\bibliographystyle{IEEEtran}
\bibliography{bib}

\end{document}